\documentclass{article}

\usepackage{microtype}
\usepackage{graphicx}
\usepackage{subfigure}
\usepackage{booktabs} % for professional tables
\usepackage{amssymb}
\usepackage{multirow}
\usepackage{caption}

\PassOptionsToPackage{hyphens}{url}
\usepackage{hyperref}

\usepackage[accepted]{mlsys2022}

\mlsystitlerunning{FastML Science Benchmarks}

\begin{document}

\twocolumn[

    %\mlsystitle{Real-Time Scientific Benchmarks for Edge Machine Learning}
    %\mlsystitle{Fast Machine Learning (Fast-ML) Benchmarks for Science}
    %\mlsystitle{Fast-ML Benchmarks for Science:\\Real-Time Scientific Benchmarks for Edge Machine Learning}
    \mlsystitle{FastML Science Benchmarks:\\Accelerating Real-Time Scientific Edge Machine Learning}

    \mlsyssetsymbol{equal}{*}

    \begin{mlsysauthorlist}
        \mlsysauthor{Javier Duarte}{equal,ucsd}
        \mlsysauthor{Nhan Tran}{equal,fnal}
        \mlsysauthor{Ben Hawks}{fnal}
        \mlsysauthor{Christian Herwig}{fnal}\\
        \mlsysauthor{Jules Muhizi}{harvard}
        \mlsysauthor{Shvetank Prakash}{harvard}
        \mlsysauthor{Vijay Janapa Reddi}{harvard}
    \end{mlsysauthorlist}

    \mlsysaffiliation{harvard}{Harvard University, Cambridge, MA, USA}
    \mlsysaffiliation{fnal}{Fermi National Accelerator Laboratory, Batavia, IL, USA}
    \mlsysaffiliation{ucsd}{University of California San Diego, La Jolla, CA, USA}

    \mlsyscorrespondingauthor{Javier Duarte}{jduarte@ucsd.edu}
    \mlsyscorrespondingauthor{Nhan Tran}{ntran@fnal.gov}
    \mlsyscorrespondingauthor{Vijay Janapa Reddi}{vj@eecs.harvard.edu}

    \mlsyskeywords{Machine Learning, Science, Real-Time, Benchmark}

    \vskip 0.3in

    \begin{abstract}
        
Applications of machine learning (ML) are growing by the day for many unique and challenging scientific applications.
However, a crucial challenge facing these applications is their need for ultra low-latency and on-detector ML capabilities.
Given the slowdown in Moore's law and Dennard scaling, coupled with the rapid advances in scientific instrumentation that is resulting in growing data rates, there is a need for ultra-fast ML at the extreme edge.
Fast ML at the edge is essential for reducing and filtering scientific data in real-time to accelerate science experimentation and enable more profound insights.
To accelerate real-time scientific edge ML hardware and software solutions, we need well-constrained benchmark tasks with enough specifications to be generically applicable and accessible.
These benchmarks can guide the design of future edge ML hardware for scientific applications capable of meeting the nanosecond and microsecond level latency requirements.
To this end, we present an initial set of scientific ML benchmarks, covering a variety of ML and embedded system techniques.
    \end{abstract}
]

\printAffiliationsAndNotice{\mlsysEqualContribution} % otherwise use the standard text.

%%%%%%%%%%%%%%%%%%%%%%%%%%%%%%%%%%%%%%%%%%
\section{Introduction}
\label{sec:intro}

\begin{figure}[th!]
    \centering
    \includegraphics[width=0.925\columnwidth]{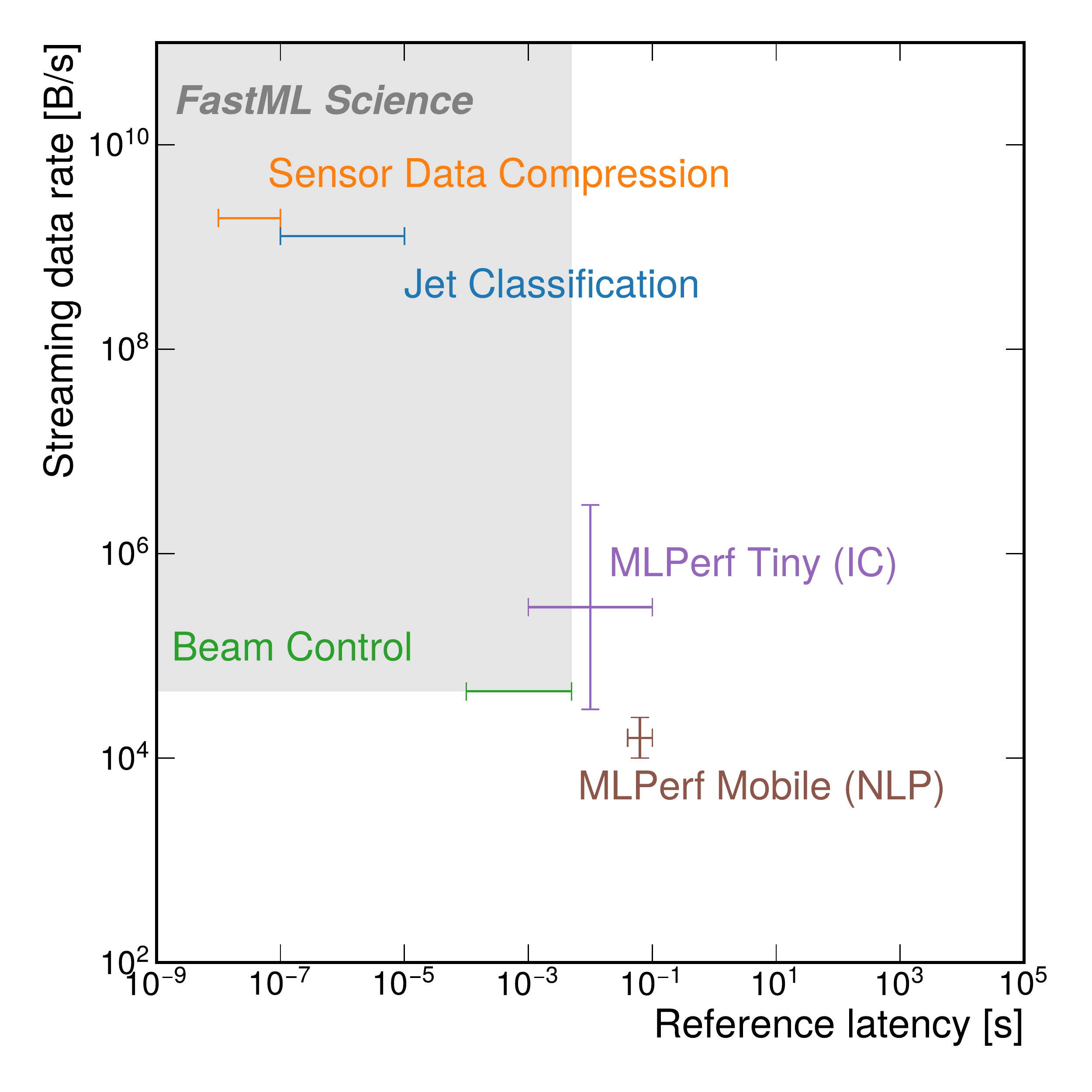}
    \caption{Reference latencies and streaming input data rates for common benchmarks and those proposed in this paper.
        The horizontal error bar represents the range of acceptable latencies for the various domains, while the vertical error bar denotes the range of streaming data rates typical for those domains.
        The real-time scientific application domain, or the FastML Science domain, produces a staggering volume of data and the inference latency requirements are orders of magnitude far more stringent than they are for more traditional consumer-facing applications and their benchmarks.}
    \label{fig:sciml}
\end{figure}

In pursuit of scientific discovery across many domains, experiments are becoming exceedingly sophisticated to probe physical systems at increasingly smaller spatial resolutions and shorter timescales.
These order of magnitude advancements have led to explosions in both data volumes and richness, leaving domain scientists to develop novel methods to handle growing data processing needs.
Figure~\ref{fig:sciml} shows the volume of data ($y$-axis) that is generated in scientific applications such as those at the CERN Large Hadron Collider (LHC) and in particle accelerator controls.
They produce tens of terabytes of data every second, as discussed below.

As scientific ecosystems snowball in their speed and scale, new data processing and reduction paradigms need to be integrated into the system-level design.
The large volume of data needs to be rapidly reduced to a sustainable level by a real-time event filter system on whether the data should be kept for further analysis or discarded.
Fortunately, this coincides with the rise of machine learning (ML), or the use of algorithms that can learn directly from data.
Recent advancements demonstrate that ML architectures based on structured deep neural networks are versatile and capable of solving a broad range of complex scientific problems.
\textit{Fast ML} is leading to rapid advancements across many different scientific domains~\cite{Carleo:2019ptp,DBLP:fastmlwhitepaper}.

However, a crucial challenge in applying ML to scientific domains is the need for ultra-fast inference speeds on the order of microseconds.
Compared to traditional ML inference latencies, the scientific application domain requires nearly 1000$\times$ faster inference performance, as shown in Figure~\ref{fig:sciml}.
Many of the consumer facing edge ML deployments like MLPerf Mobile and Tiny are at least an order of magnitude off from the scientific application latency requirements.

% The proliferation of large datasets like ImageNet~\cite{imagenet}, computing, and DL software has led to the exploration of many different DL approaches each with their own advantages. 

Moreover, while ML can power scientific advances that can lead to future paradigm shifts in a broad range of scientific domains, including particle physics, accelerator physics, plasma physics, astronomy, neuroscience, chemistry, material science, and biomedical engineering, each application brings its own domain-specific knowledge and application-specific constraints.
Furthermore, there can be a broad range of different challenges even within a given domain.
This leaves us with two primary types of challenges.
The first is how can we define generically applicable \textit{ML benchmarks tasks} for bespoke domain problems which can attract interest from a broad community of system and ML experts?
The second is how can we design benchmark tasks to satisfy \textit{challenging} system-level scientific requirements that can, at its core, have common elements which overlap with a number of system architectures?
To address these issues, we introduce a domain-specific benchmark for fast ML science.

In this paper, we address the aforementioned challenges by defining an initial set of scientific ``edge'' ML benchmarks.
These benchmarks strike a balance between a universality common to many scientific ML applications while also being uniquely challenging to push designers to further the state-of-the-art ML and system architecture beyond traditional industrial applications.
We choose our benchmark tasks to cover the three different areas of ML, also we select domain tasks that, while seeming quite specific, can be generically mapped onto other scientific applications.
In generic terms, the initial benchmark tasks are the following:
\begin{itemize}
    \item \textbf{Supervised learning for physics event triggering}: feature extraction of point cloud inputs for rare data classification and filtering.
    \item \textbf{Unsupervised learning for lossy compression of particle detector data}: near-sensor image data compression for custom sensor geometries using image similarity metrics based on energy distribution profiles.
    \item \textbf{Reinforcement learning for accelerator beam control}: optimal real-time system controls based on time-series data from multiple sensors.
\end{itemize}

% In each of these scenarios, we are working at the edge, very near to the data source.  
% Often, the input data has just been digitized,\footnote{In even more extreme scenarios in the future, we may consider analog inputs.} and is therefore quantized with a custom integer precision.  
% As is often the case in experimental design, we will then lay out system constraints which will provide guide rails for the benchmark tasks.  
% Solutions for the scientific ML edge benchmarks must then satisfy these requirements and then performance will be measured.  
% Each benchmark will have specific constraints. 
% They are all latency-bound, with latency requirements ranging from hundreds of nanoseconds to milliseconds, and in some cases the algorithm latency and the data arrival frequency (pipeline interval) will not be the same.  
% In one application, we have requirements on area and power as well.   

With this first suite of benchmarks, we provide suitable coverage for scientific tasks that span a range of different ML areas and include unique, meaningful, and challenging system-level requirements that go beyond those from industry.
Future work would include finding areas not covered by these tasks and broadening the benchmarks.

% One of the overarching attractions of ML is not only the power of the algorithms but also the generality of the approach. But this matrixed high dimensionality of domains, tasks, and systems presents a considerable barrier. 
% It is one of the primary reasons that progress in scientific disciplines is stove-piped and limited with one-off engagement from system-level experts. 
% In a recent community white paper~\cite{DBLP:fastmlwhitepaper}, an early attempt is made to call out opportunities for scientific advancement with real-time ML across a wide array of domains. 
% This includes identifying areas of commonality and overlap where advancements can broadly serve multiple domains. 
%%%%%%%%%%%%%%%%%%%%%%%%%%%%%%%%%%%%%%%%%%
\section{Related Work}

Table~\ref{table:MLBenchmarkComparison} compares characteristics of some of the existing science-related benchmarks and initiatives to the real-time scientific edge ML benchmarks we present in this work.
Most of the prior work included in the table has been formalized as an official benchmark, but a couple of related initiatives remain as standalone datasets or competitions.

\renewcommand{\arraystretch}{1.2}
\begin{table*}[t!]
  \centering
  \begin{tabular}{|c|c|c|c|c|}
    \hline
                                                          & Formalized    &
    Scientific                                            &
    Edge                                                  &
    Real-Time                                                                                                             \\
                                                          & Benchmark     &
    Workload(s)                                           &
    Computing                                             &
    Constraints                                                                                                           \\
    \hline\hline\hline
    \bf FastML Science Benchmarks (this work)             & \bf\checkmark & \bf\checkmark & \bf\checkmark & \bf\checkmark \\
    \hline\hline
    \hline
    SciMLBench~\cite{thiyagalingam2021scientific}         & \bf\checkmark & \bf\checkmark & \bf\checkmark & $\times$      \\
    \hline
    LHC New Physics Dataset~\cite{govorkova2021lhc}       & $\times$      & \bf\checkmark & \bf\checkmark & \bf\checkmark \\
    \hline
    MLPerf HPC~\cite{farrell2021mlperf}                   & \bf\checkmark & \bf\checkmark & $\times$      & $\times$      \\
    \hline
    BenchCounil AIBench HPC~\cite{benchcouncil}           & \bf\checkmark & \bf\checkmark & $\times$      & $\times$      \\
    \hline
    MLCommons Science~\cite{mlcommonsscience}             & \bf\checkmark & \bf\checkmark & $\times$      & $\times$      \\
    \hline
    ITU Modulation Classification~\cite{ituML5Gchallenge} & $\times$      & $\times$      & \bf\checkmark & \bf\checkmark \\
    \hline
  \end{tabular}
  \caption{Comparison of existing machine learning benchmarks and related initiatives.}
  \vspace{-10pt}
  \label{table:MLBenchmarkComparison}
\end{table*}

Among existing AI benchmarks, the community-driven MLPerf benchmarks from MLCommons~\cite{mattson2020mlperf} are well-established.
The benchmarks are run under predefined conditions and evaluate the performance of training and inference for hardware, software, and services.
MLPerf regularly conducts new tests and adds new workloads to adapt to the latest industry trends and state of the art in AI across various domains including high performance computing (HPC)~\cite{farrell2021mlperf}, datacenter~\cite{inferencedatacenter2021}, edge~\cite{inferenceedge2021}, mobile~\cite{reddi2020mlperf}, and tiny~\cite{banbury2021mlperf}.
Additionally, BenchCouncil AIBench is a comprehensive AI benchmark suite including AI Scenario, Training, Inference, Micro, and Synthetic Benchmarks across datacenter, HPC, IoT and edge~\cite{benchcouncil}.
Other benchmarks have also been developed by academia and industry.
Additional examples of prior art and initiatives include AI Benchmark ~\cite{ignatov2019ai}, EEMBC MLMark~\cite{torelli2019measuring}, AIMatrix~\cite{aimatrix}, AIXPRT~\cite{aixprt}, DeepBench~\cite{deepbench}, TBD~\cite{zhu2018benchmarking}, Fathom~\cite{adolf2016fathom}, RLBench~\cite{james2020rlbench}, and DAWNBench~\cite{coleman2017dawnbench}.

However, scientific applications (i.e. cosmology, particle physics, biology, clean energy, etc.) are innately distinct from traditional industrial applications with respect to the type and volume of data and the resulting model complexity~\cite{farrell2021mlperf}.
The MLCommons Science Working Group~\cite{mlcommonsscience} has a suite of benchmarks that focus on such scientific workloads including application examples across several domains such as climate, materials, medicine, and earthquakes.
SciMLBench~\cite{thiyagalingam2021scientific} from the Rutherford Appleton Laboratory is another benchmark suite specifically focused on \textit{scientific machine learning} and aimed towards the ``AI for Science" domain.
The suite currently contains three benchmarks that represent problems taken from the material and environmental sciences.
MLPerf HPC and AIBench HPCAI500 are two more benchmarks that include scientific workloads.
In general, HPC is being leveraged by the scientific community for accelerating scientific insights and discovery.
MLPerf HPC aims to systematically understand how scientific  applications perform on diverse supercomputers, focusing on the time to train for three representative scientific machine learning applications with massive datasets (i.e. cosmology, extreme weather analytics, and molecular dynamics).
Similarly, AIBench HPCAI500 also includes a benchmark on extreme weather analytics.
However, all of these existing benchmarks fail to capture the unique constraints and demands required by scientific ``edge" computing in which the processing must occur near the data source in \textit{real-time}.

The LHC physics dataset for new physics detection~\cite{govorkova2021lhc} is the only dataset to our knowledge that retains such characteristics, requiring unsupervised detection at 40\,MHz.
However, a related dataset for charged particle tracking at the LHC, the throughput phase of the TrackML Challenge~\cite{Amrouche:2021tio} emphasized balancing the accuracy of the solution with the speed of inference.
Similarly, one of the tasks in the International Telecommunication Union's (ITU) ML in 5G Challenge~\cite{ituML5Gchallenge} requires ultra-low latency for a different application, namely modulation classification in communication networks, which displays the generalizability of real-time scientific edge ML benchmarks.
Benchmarking workloads of this nature can ultimately be a catalyst in enabling a diverse range of new solutions and applications.

%%%%%%%%%%%%%%%%%%%%%%%%%%%%%%%%%%%%%%%%%%
\section{Benchmarks}

In this section, we first introduce our benchmark design philosophy to establish a suite of real-time scientific benchmarks for edge machine learning.
Next, we introduce the tasks, models, datasets, and associated metrics that define this initial suite.
These benchmarks are in the initial stages and the task list will continue to evolve over time.

\subsection{Benchmark Design Philosophy}

Our benchmarks tasks cover supervised learning, unsupervised learning, and reinforcement learning and map generically onto other scientific applications.
The benchmarks are summarized in Table~\ref{table:summary}, which we explain in more detail in the subsequent sections.

In each benchmark task, we are working at the edge, very close to the data source.
For each task, the input data has been digitized\footnote{In even more extreme scenarios in the future, we may consider analog inputs.}, but the analog signals from sensors are digitized via custom ADCs that have a custom precision that is tailored for the sensor technology. Scientific instrument sensors often may have a non-standard precision and further, may deal with inputs of varying quantization.
As is often the case in experimental design, we lay out system constraints that will serve as guide rails for the various benchmark tasks.

Solutions for the scientific ML edge benchmarks must satisfy the given requirements, and then the algorithm performance is measured by a task-specific metric.
Each benchmark has specific system-level constraints.
They are all latency-bound, with latency requirements ranging from hundreds of nanoseconds to milliseconds.
In some cases, the algorithm latency and data arrival frequency (pipeline interval) will not be the same.
In one application, we also have requirements for area and power.

In order to standardize measurement an implementation's performance of a given benchmark against a potentially diverse set of other hardware and technologies, a number of related metrics should be captured and presented to more accurately characterize a given aspect of a system.
For example, when measuring the latency of an implementation, both the end-to-end latency and the initiation interval, or the time required before a system is ready to accept an input sample after it's just received one, should be measured and presented as part of the results.
To accurately support and measure various comprehensive metrics, careful implementation of a ``host'' system, which is responsible for administering a benchmark, must also taken into account.
Continuing the example of measuring latency of an implementation, the timing and method which inputs are presented to the system under test must be capable of strictly obeying the benchmark's pipeline interval.
A framework for performing such measurements on varied hardware which we can adopt is based on the MLPerf Tiny benchmark~\cite{banbury2021mlperf}.

\subsection{Supervised Learning: Jet Classification}

At the LHC, proton bunches collide at an extreme frequency of 40\,MHz, and data rates at the two multipurpose high energy particle physics experiments, CMS~\cite{CMS} and ATLAS~\cite{ATLAS}, are of the order of hundreds of terabytes per second.
With such high data rates, the task of real-time processing to filter events to reduce data rates to manageable levels for offline processing is called triggering.
The first level of the trigger at CMS~\cite{CMSL1T} and ATLAS~\cite{ATLASL1T} is performed in FPGAs in custom electronics platforms and have latency requirements at the microsecond scale; to put this into proper perspective, it is worth noting that the MLPerf inference tasks' latency typically ranges between 10 to 100s of milliseconds.

\begin{figure}[t]
    \centering
    \includegraphics[width=0.45\textwidth]{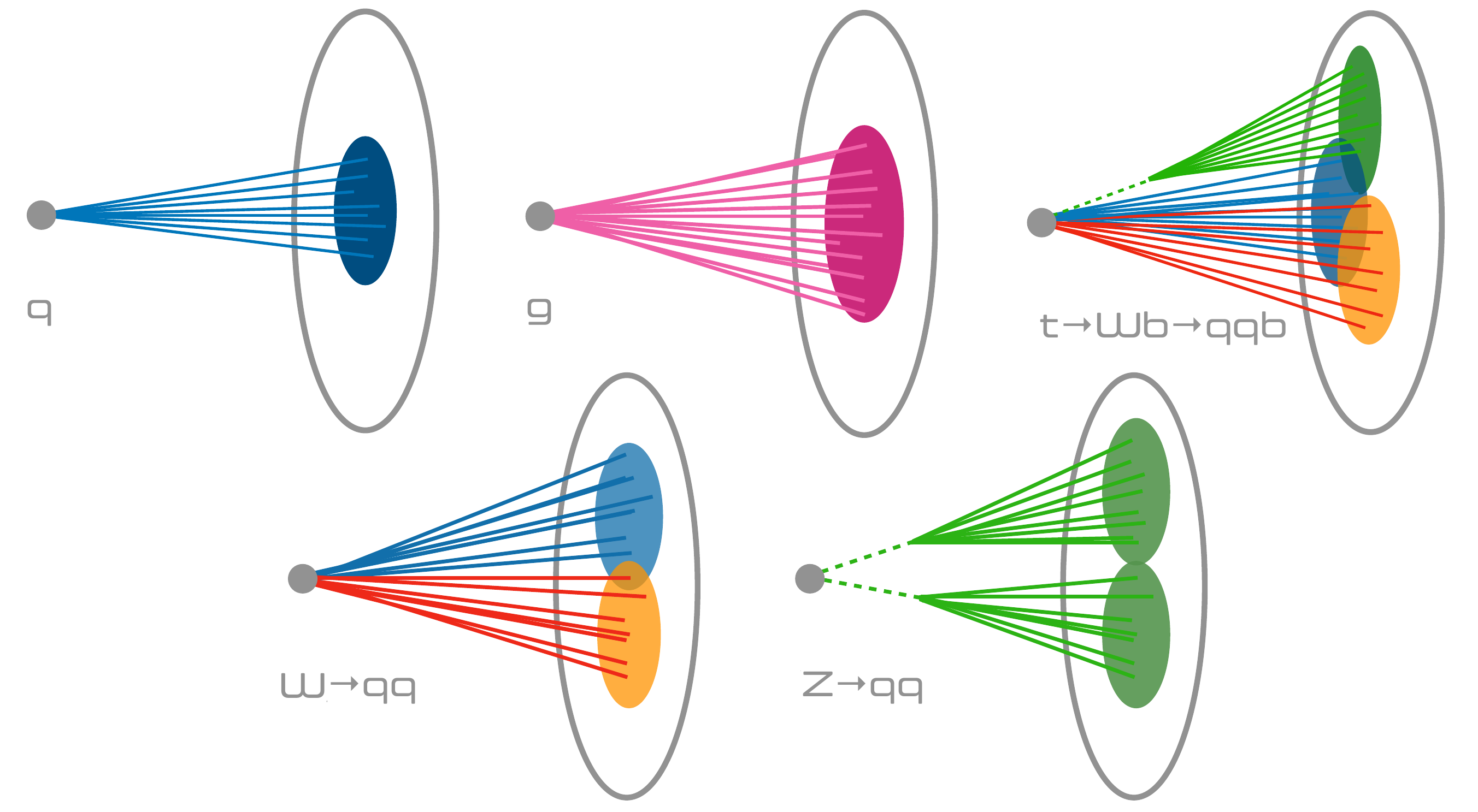}
    \caption{The five jet type classes---light quark (q), gluon (g), top quark (t), W, and Z boson---for the supervised learning benchmark task.
        Figure adapted from \cite{Moreno:2019bmu}.}
    \label{fig:Jets}
\end{figure}

While this task is specific to particle physics, filtering data to find the most interesting subset in real-time to reduce data transmission and volumes in big data scientific experiments is a very common challenge.
In this scenario, we are looking for proton collisions with interesting energetic radiation patterns called \textit{jets}.
In particular, we would like to identify rare jet signatures originating from the W or Z boson or the top quark (t) in contrast to more common signatures from the lighter quarks (q) or gluons (g).
Due to high fidelity simulations in particle physics experiments, this becomes a supervised multi-classification task.
A schematic illustration of the different types of jets is shown in Fig.~\ref{fig:Jets}.

\paragraph{Dataset}

We allow for two types of input features.
Several previous related studies used expert features in a dense, fully-connected network topology which we will describe in the following section as our baseline model.
There are 16 expert-designed features, traditionally standardized with a standard scalar and represented with fixed-point precision with 16 total bits and six integer bits.
In realistic systems, the classification task would also require the computation of those expert features, but we do not need that here.
However, to that end, we make available a more challenging set of point cloud inputs which are the 100 most energetic particles in the jet (with zero padding if there are less than 100 particles)~\cite{pierini_maurizio_2020_3602254}.
Each particle is assumed to be massless and has a feature set size of 3 corresponding to the momentum vector.
Models built from these input particles can be more performant than the expert features but could be more computationally expensive.
Given that the range of $p_x$, $p_y$, and $p_z$ for an individual particle is approximately contained within $\pm$2\,TeV, we adopt an input integer quantization scheme of 16 total bits with a least significant bit corresponding to 0.0625\,GeV and a total range of $\pm$2048\,GeV.

\paragraph{Performance Metrics}

There are many metrics used in the literature for this benchmark model.
The two we will focus on are (1) classification accuracy and (2) FPR at TPR of 50\% for the signal being Z jet~\cite{10.3389/frai.2021.676564}.

\paragraph{Real-time System Constraints}

In the planned upgrade of the CMS trigger, the global correlator trigger would contain jet tagging algorithms like the one described.
This system features a time-multiplexed design, such that information from 6 consecutive proton beam crossings, which occur even 25\,ns, is processed simultaneously.
For this system, a latency of no more than 1\,$\mu$s and the ability to accept new inputs every 150\,ns is required~\cite{CMSP2L1T}.

\paragraph{Baseline Model and Implementation}
For our baseline model and performance, we consider the 16-input dense, fully-connected architecture first presented in Ref.~\cite{Duarte:2018ite} and based on expert inputs.
Because we consider the model with expert inputs, a fully-connected architecture is selected, but other neural network architectures may be more optimal, particularly if particle level inputs are used.

The baseline model has an accuracy of 74.8\% and the FPR at TPR of 50\% metric for the baseline model is 0.00129.
The baseline model implementation with the architecture consists of $16 \rightarrow 64 \rightarrow 32 \rightarrow 32 \rightarrow 5$ MLP with quantization-aware training (QAT) to a homogeneous precision for the weights and biases of 6-bits as presented in Ref.~\cite{Coelho:2020zfu}.
This model is synthesized for a Virtex Ultrascale+ 9P FPGA with the following resource usage: 124\,DSPs, 39782\,LUTs, and 8128\,FFs.
The latency is 55\,ns at a 5\,ns clock with a pipeline interval of 1 clock cycle (5\,ns).

%%%%%%%%%%%%%%%%%%%%%%%%%%%%%%%%%%%
\subsection{Unsupervised Learning: Irregular Sensor Data Compression}

Processing of LHC events requires two parallel streams of detector data to be transmitted from on-detector readout chips.
The first is complete event data from all sensing elements, transmitted at the trigger accept rate of 100\,kHz, while the second is a compressed representation of the same data at the full 40\,MHz collision rate.
This second, lightweight representation is the basis for trigger decisions, with a compression factor of $\mathcal{O}$(400), ensuring parity between the bandwidth of each stream.
The critical task is to achieve this reduction in $\mathcal{O}$(100\,ns) with minimal impact on downstream physics algorithms.

\begin{figure}[t]
    \centering
    \includegraphics[width=0.49\textwidth]{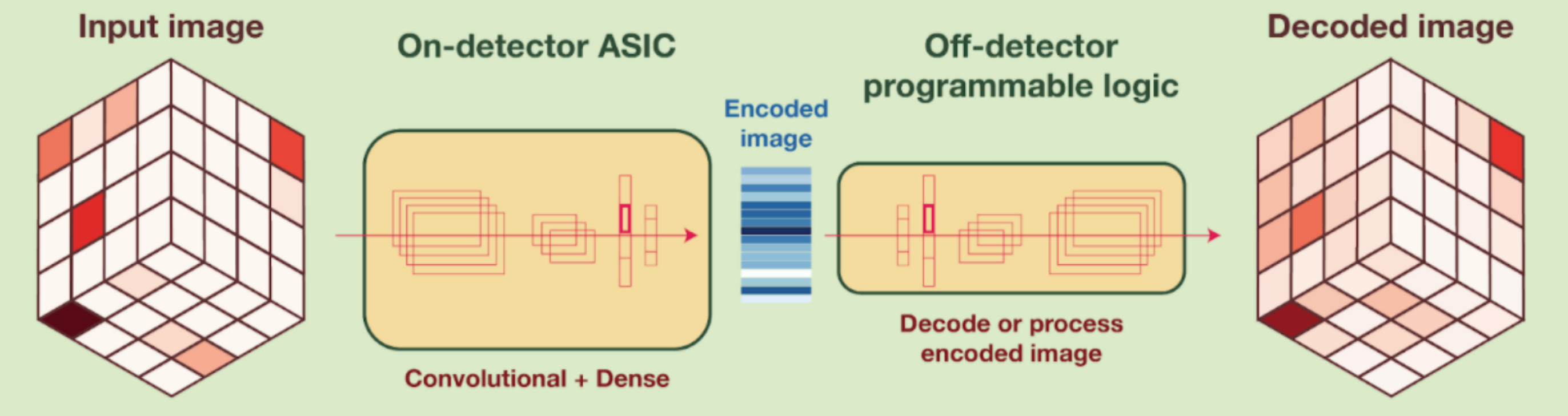}
    \vspace{-10pt}
    \caption{Schematic of the compress-transmit-decompress pipeline for the HGCal trigger data stream.}
    \label{fig:ASIC}
\end{figure}

This task is common to many detector systems and directly relevant to a host of on-device compression tasks for complex sensor data.
The scenario considers the CMS high-granularity endcap calorimeter (HGCal), comprised of 6M channels, each capturing a 5d (position, energy and time) image of showering high-energy particles.
HGCal is comprised of layers of hexagonal arrays, with a single particle depositing energy into hundreds of individual sensors.
Data from each hexagonal array is compressed in an application-specific integrated circuit (ASIC), with the encoded representation transmitted off-detector and subsequently used to recover the initial detector image as in Figure~\ref{fig:ASIC}.
A data-driven approach is required to tune the parameters of the compress and decompress algorithms to minimize differences between the original and decoded images.

\paragraph{Dataset}

Compression of the trigger data is accomplished in multiple steps.
First, nearest-neighbors within each hexagonal array are aggregated to form a set of 48 ``trigger cells'' with energies represented as a custom 7b float.
These 48$\times$7b floating-point inputs are converted to 22b integers, summed, and finally normalized to obtain 8b fixed-point trigger cell data.
The sum is preserved as a 9b float for transmission.
The benchmark task begins from this point, which must capture this $48 \times 8=384$b representation in a budget of either 144 or 48 bits, for the moderate---and extreme---compression variants, respectively.

\paragraph{Real-time System Constraints}

The encoder must accept new inputs at 40\,MHz input rate and complete processing within 100\,ns.
Furthermore, we must consider protection against single-event effects due to the high-radiation environment of the on-detector readout.
As a rough first order guideline, on a low-power CMOS 65\,nm technology node, the algorithm area must not be greater than 4\,mm$^2$ while drawing less than 100\,mW.

\paragraph{Performance Metrics}

Performance is assessed by directly comparing the individual energies of each decoded array of hexagonal sensors (i.e. 48 trigger cells) to the original image of normalized inputs.
The energy mover's distance~\cite{emd} (EMD) is used to compare the reconstructed radiation patterns, giving smaller penalties for misreconstructed energies that are close-by to the original deposit.

%(\textcolor{blue}{CH: any words here on a decoder implementation?})(\textcolor{red}{NT: I added area constraint}) 

\paragraph{Baseline Model}

A convolutional neutral network (CNN) autoencoder architecture is used to perform the compress-and-decompress task because the sensor data is image-like, though the hexagonal geometry provides a unique challenge.
Normalized sensor data is re-arranged and fed to a CNN with one convolutional and one dense layer with 6b weights, leading to a maximum of 16$\times$9b outputs, saturating the 144b bandwidth~\cite{DiGuglielmo:2021ide}.
Data is decompressed by a second CNN with inverted architecture, whose outputs are multiplied by the energy sum to recover the original input image.
%The EMD metric mean is \textcolor{red}{XX.XX}. 

The compression logic is implemented in an ``encoder ASIC'' using a low-power CMOS process with 65\,nm feature size.
The total latency for this circuit is 25\,ns and is estimated to draw 60\,mW in simulation.

%%%%%%%%%%%%%%%%%%%%%%%%%%%%%%%%%%%%%%%%%%%%
\renewcommand{\arraystretch}{1.2}
\begin{table*}[t]
    \centering
    \resizebox{.95\textwidth}{!}{
        \begin{tabular}{|c|c|c|c|c|c|c|}
            \hline
            \multirow{2}{*}{Type}       &
            \multirow{2}{*}{Benchmark}  &
            Input                       &
            Pipeline                    &
            Real-time                   &
            \multirow{2}{*}{Misc. Req.} &
            Baseline Model                                                                                                              \\
                                        &                         & Precision & Rate    & Latency   &                      & Parameters \\
            \hline\hline
            Supervised Learning         & Jet Classification      & 16b       & 150\,ns & 1\,$\mu$s & -                    & 4,389      \\
            \hline
            Unsupervised Learning       & Sensor Data Compression & 9b        & 25\,ns  & 100\,ns   & area, power (65\,nm) & 2,288      \\
            \hline
            Reinforcement Learning      & Beam Control            & 32b       & 5\,ms   & 5\,ms     & -                    & 34,695     \\
            \hline
        \end{tabular}
    }
    \captionsetup{justification=centering}
    \caption{Summary of constraints for three benchmark tasks and number of parameters for the benchmark baseline models.}
    \vspace{-10pt}
    \label{table:summary}
\end{table*}
%%%%%%%%%%%%%%%%%%%%%%%%%%%%%%%%%%%%%%%%%%%%

\subsection{Reinforcement Learning: Beam Control}

Intense and energetic particle beams are used for various applications, from materials discovery to studying nuclear matter to fundamental particle physics, and even cancer therapy.
Controlling precise particle beams, such as those at the Department of Energy User Facilities, requires intelligent algorithms running at the edge in low-latency, real-time systems to steer particles traversing miles of beamline at nearly the speed of light.

\begin{figure}[t]
    \centering
    \includegraphics[width=0.45\textwidth]{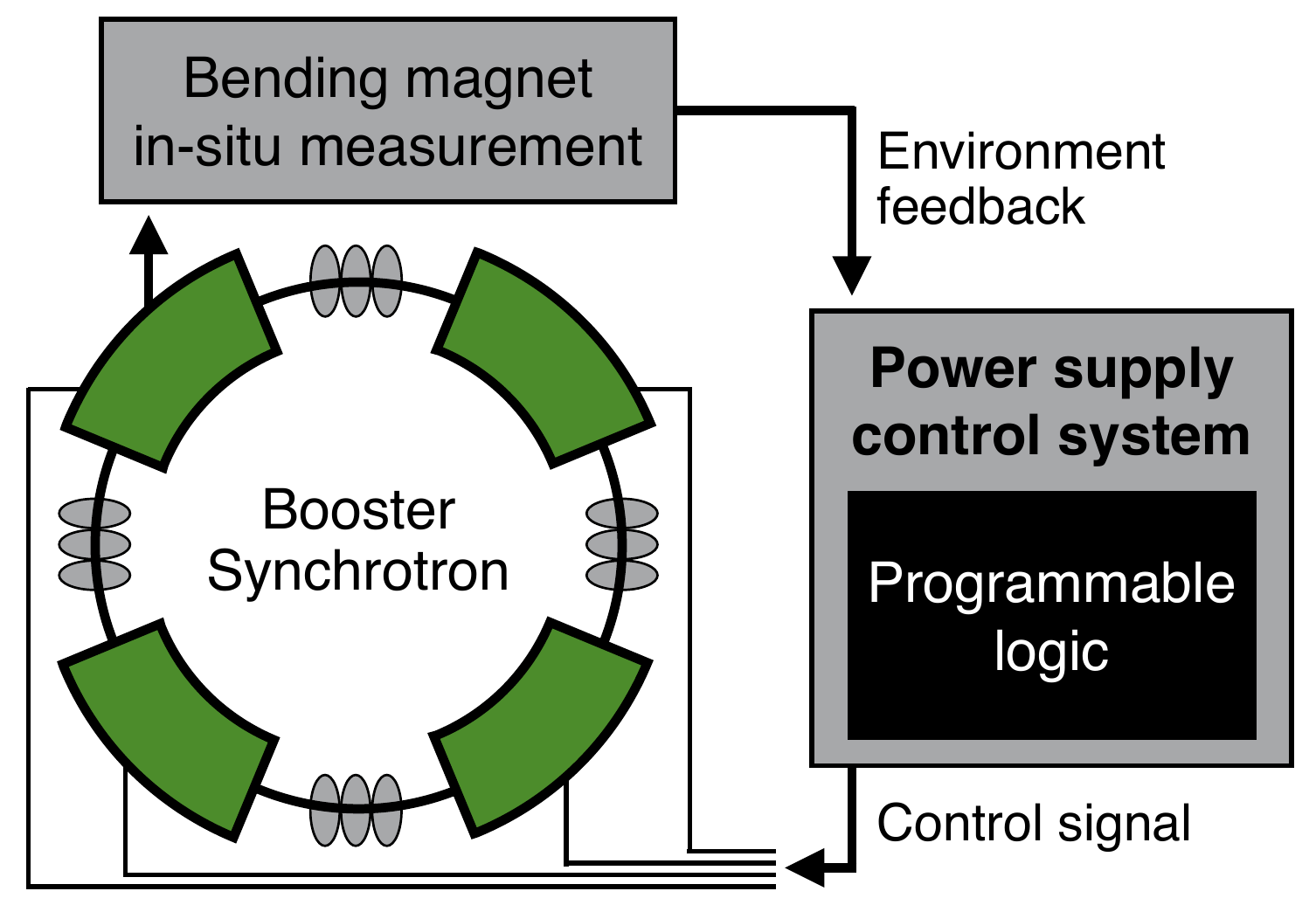}
    \caption{Synchrotron magnet power supply control system for the Fermilab Booster Ring, adapted from~\cite{StJohn:2020bpk}}
    \label{fig:booster}
\end{figure}

A dataset has been developed~\cite{kafkes2021boostr} for studying how to control the bending magnet ramping rate of power supplies~\cite{StJohn:2020bpk} in the rapidly cycling Booster synchrotron ring~\cite{booster} at the Fermilab Accelerator Complex.
This is illustrated diagrammatically in Figure~\ref{fig:booster}.
The power supply control signals are provided at 15\,Hz.
This beam controls application can be framed as a reinforcement learning benchmark task.
Because an accurate and reliable simulation of the synchrotron is not possible from first principles, a ``virtual'' accelerator complex surrogate model has been developed to emulate the actual physical system.
This surrogate model will serve as the environment with which our reinforcement learning benchmark interacts.

\paragraph{Dataset}

A Booster synchrotron power supply regulation dataset provides cycle-by-cycle time series of readings and settings from the most relevant devices available in the Fermilab control system.
This data was drawn from the time series of a select subset of the roughly 200,000 entries that populate the device database of the accelerator control network.
Data was sampled at 15\,Hz for 54 devices pertaining to the system's regulation. Because of how data is transmitted and communicated, inputs are 32-bit floating-point numbers, but the sensor source's precision is, in many cases, less.

\paragraph{Real-time System Constraints}
The Booster ramping cycle rate is 15\,Hz, which sets the control loop's time scale.
We define the algorithm latency requirement as 5\,ms for this benchmark due to data movement latency.

\paragraph{Performance Metrics}
The primary performance metric in this reference benchmark is the reward, $R$, defined as the negative of the error with respect to the reference expected current in the Booster, $R = -|\Delta I_\mathrm{min}|$.

\paragraph{Baseline Model(s)}
There are two models involved in this benchmark task: (1) the surrogate model for the Booster accelerator and (2) the online agent, which is correcting the reference magnet power supplies in real-time.
The surrogate model is fixed in this benchmark task and plays the role of the environment in this reinforcement learning task.
The long short-term memory (LSTM)~\cite{lstm_paper} surrogate model inputs are the previous 150-time steps of the top 5 causal variables---variables related to the synchrotron and downstream accelerator currents and current errors concerning reference.
The model has approximately 1.5 million parameters.

The benchmark online agent running in the Arria10 system-on-chip (SoC) is a multilayer perceptron taking the five input parameters, has three hidden layers $(128,128,128)$ and approximately 35,000 parameters.
The architecture is a fully-connected neural network because the 5 inputs are already selected expert variables.
The deep Q-network~\cite{mnih2013playing,Mnih2015HumanlevelCT} has 7 discrete outputs and maximizes the reward, $R$, defined above.
The reward metric is measured as a function of RL episode and is presented in Fig.~7 (bottom) of ~\cite{StJohn:2020bpk}.
% Describe the surrogate model and the baseline agent model.  

The benchmark model weights and biases are quantized to 20 total bits in a fixed-point representation in hardware.
The lowest latency implementation of the model is implemented for an Intel Arria10 SoC with a resource usage of 53\,DSPs, 238\,BRAMs, 672\,MLABs, 43.3\,kALMs, 92.6\,kFFs.
The algorithm has a latency of 3.9\,$\mu$s.

\section{Discussion and Outlook}
This position paper highlights both the need and challenges for developing machine learning (ML) benchmarks for edge applications in science.
Given the demise of Moore's law and Dennard scaling~\cite{dennard,breakdown} and advances in scientific instrumentation resulting in rapidly growing data rates, edge computing is becoming exceedingly crucial for reducing and filtering scientific data in real-time to accelerate science experimentation and enable more profound insights.
There are challenges in building well-constrained benchmark tasks with enough specification to be generically applicable and accessible simultaneously.
However, we can use these edge applications in extreme data processing environments to advance many scientific domains and enable the development of state-of-the-art tools.

In devising scientific ML edge benchmarks tasks, we aim to cover machine learning and embedded system techniques.
We choose three applications that span supervised, unsupervised, and reinforcement learning.
The system constraints also span a wide range of latency requirements---from hundreds of nanoseconds to milliseconds where technologies vary from ASIC to FPGA to more relaxed constraints with the freedom to choose system architecture.
Finally, there is also a variation in input data representations from expert-level inputs to image data, point cloud data, and time-series data. 
A summary of the proposed scientific edge ML benchmarks are presented in Table~\ref{table:summary} including the input, latency, and pipeline interval constraints of the benchmark applications. 
We also provide a prototype code repository for these benchmarks~\cite{jules_muhizi_2022_5866587}.

\section{Conclusion}

In this work, we present an initial set of scientific machine learning benchmarks that are specifically geared towards real-time scientific edge machine learning needs.
Our goal is to first and foremost identify and provide a suite of datasets and a code repository for these various edge ML tasks.
We will collect and document different solutions and, based on interest, extend this work. 
In particular, there is potential to provide more domain applications such as for astronomy, neuroscience, and microscopy that could attract a broader set of use-cases.

\section*{Acknowledgments}

JD is supported by the US Department of Energy (DOE) Award Nos. DE-SC0021187, DE-SC0021396 and the National Science Foundation (NSF) A3D3 Institute under Cooperative Agreement OAC-2117997.
BH, CH, JM, and NT are supported by Fermi Research Alliance, LLC under Contract No. DE-AC02-07CH11359 with the U.S. Department of Energy (DOE), Office of Science, Office of High Energy Physics, the DOE Early Career Research program under Award No. DE-0000247070, and the DOE, Office of Science, Office of Advanced Scientific Computing Research under Award No. DE-FOA-0002501.

% are suppored by Fermi Research Alliance, LLC under Contract No. DE-AC02-07CH11359 with the U.S. Department of Energy, Office of Science, Office of High Energy Physics.

%%%%%%%%%%%%%%%%%%%%%%%%%%%%%%%%%%%%%%%%
% \clearpage
% \appendix
% \section{Appendix? What do we want in a repo (Jules, ...)}

% Could we have a demo repo of the particle jet classification problem? 

% \begin{itemize}
%     \item dataset (on zenodo already)
%     \item data loader - both for expert features and top N particles
%     \item script for training (PTQ, original), maybe a QAT version; MLP on expert features, conv1D on particle level features
%     \item script to calculate the performance metrics - Accuracy, AUC (avg), FPR at TPR (x5, average?, pick one?) 
%     \item resource metrics (optional), script to calculate: BOPs
% \end{itemize}

% \clearpage

% \clearpage 
\bibliography{references}
\bibliographystyle{mlsys2022}

%%%%%%%%%%%%%%%%%%%%%%%%%%%%%%%%%%%%%%%%%%%%%%%%%%%%%%%%%%%%

\end{document}